# Artificial Agency and Large Language Models

Maud van Lier

Philosophy Department
University of Konstanz
Universitätsstraße 10,
78467 Konstanz, Germany
*maud.van-lier@uni-konstanz.de*

Gorka Muñoz-Gil

Institute of Theoretical Physics,
University of Innsbruck
Innrain 52,
6020 Innsbruck, Austria
*gorka.munoz-gil@uibk.ac.at*

The arrival of Large Language Models (LLMs) has stirred up philosophical debates about the possibility of realizing agency in an artificial manner. In this work we contribute to the debate by presenting a theoretical model that can be used as a threshold conception for artificial agents. The model defines agents as systems whose actions and goals are always influenced by a dynamic framework of factors that consists of the agent's accessible history, its adaptive repertoire and its external environment. This framework, in turn, is influenced by the actions that the agent takes and the goals that it forms. We show with the help of the model that state-of-the-art LLMs are not agents yet, but that there are elements to them that suggest a way forward. The paper argues that a combination of the agent architecture presented in Park *et al.* (2023) together with the use of modules like the Coscientist in Boiko *et al.* (2023) could potentially be a way to realize agency in an artificial manner. We end the paper by reflecting on the obstacles one might face in building such an artificial agent and by presenting possible directions for future research.

## 1. INTRODUCTION

It is a common practice in computer science and artificial intelligence (AI) to refer to a certain group of artificial systems as 'agents'. A general denominator of such systems is that they can interact with the environment that they are in, meaning that they can perceive their environment through sensors and act on it through actuators (see Russell et al., 2022, p. 54). Until recently, this has been about the only feature that these systems share in common with the kind of entities that philosophers have been referring to as agents. This is because in the philosophies of mind, action, and agency, agents are systems that are autonomous in the sense that things can be *up to them*. So, where both an electric door and a human can be said to interact with their environment, it is only the human that seems to have a real say in *how* s/he interacts with this environment. This does not of course mean that philosophers hold that agents can do just *anything*. If we take a cat as an example, then we would say that it can be up to the cat whether it scratches the couch or pushes the glass off the table, but that it is not similarly up to it whether it sheds hair or eats catnip. Agents are physical systems embedded in an environment and what can be up to them will depend on a number of factors, like their physical embodiment, their experiences over time, and what options or restrictions their environment affords them. These factors, in turn, are influenced by what such an agent --- as an individual system of a particular physicality --- experiences over time. An agent, in philosophy, is thus a dynamic system that things can be up to, where what can be up to this system is influenced by a number of factors that all relate to the way that this particular system has been embedded in its environment over time.

Such an agent is quite different from the kind of systems that computer scientists have been referring to as agents. The 'acts' of these systems are not affected by any factors that directly relate to those systems as particular individuals. Where the mars-rover or a system like AlphaFold (Jumper *et al.*, 2021) can do quite astonishing things *by themselves*, what they end up doing is still prescribed by us humans. After all, it does not seem to be up to the mars-rover *if* it will explore, nor *how* it will do so. Similarly, it is not up to AlphaFold *whether* it is going to predict protein-shapes, or *what* method it will use to learn how to predict these shapes. Given that things are not similarly up to artificial systems as they are to humans and higher-order animals, most philosophers have not felt the need to draw any serious connection between the artificial 'agents' in computer science and AI and the entities that they have referred to as agents.

This attitude has changed somewhat with the arrival of Large Language Models (LLMs) like Chat-GPT. Symons and Abumusab (2024), for instance, have argued that threshold conceptions of agency typical of philosophy (those that provide necessary and sufficient conditions for agency) hinder a proper understanding of the way that LLMs are affecting our social systems. They claim that "understanding aspects of agency and recognizing that they can be productively studied in terms of dimensions and degrees are both realistic and more methodologically fruitful in the ethics of AI than traditional threshold accounts" (p. 4). We agree with Symons and Abumusab that current LLMs require the attention of philosophers and that traditional threshold accounts in philosophy are inadequate to evaluate whether we can attribute agency to these artificial systems. However, we believe that the reason that these traditional accounts are inadequate for such an evaluation is because they take human agency as the standard model for *any* form of agency. In Swanepoel (2021), for example, the author holds that agents are entities that can do things with *intent,* where intentionality requires mental states like beliefs and desires. Such a threshold conception of agency does not only exclude artificial systems, but most animals as well, and is thus not a proper representation of agents in general. Still, this does not mean that we should get rid of threshold conceptions of agency altogether as Symons and Abumusab suggest. This is because there is still a categorical difference between a system that displays some agent-like qualities and one that *has* agency.

An attribution of agency to a system changes the way that we behave towards such a system. We can expect certain things from agents that we cannot expect from non-agents, and we adjust our own behavior based on these expectations. Having a cat that has developed a liking to pushing things off surfaces causes us to stop leaving mugs on tables when we are not there to watch the cat. Where we do not know for certain that the cat will always push things off the table, we know that we can expect this kind of behavior from it. Even though we can thus not fully predict the behavior of agents, we are quite capable of anticipating for each agent what range of behaviors they are most likely to exhibit in a particular situation, and we can adjust our own behavior accordingly.

Our interactions with current LLMs take place at a very sophisticated level (see Xi et al. (2023) for a comprehensive overview) and it is therefore quite likely that, if they were to be actual agents, we would interact with them as we do with *human* agents. This means giving them responsibilities that we would normally only trust unto beings like us: rational and moral agents. This is quite different from how we currently treat artificial systems or machines. We rely on machines, but do not trust them. We do not expect them to understand why something is wrong, or to have our well-being at heart. If LLMs are not only agents, but also agents like us, then we can integrate them in our society as we would ourselves --- as an active participant that can be trusted to be left to its own devices. If, however, current LLMs are rather systems that exhibit agent-like qualities, then we should still have fail-safes in place.[1] We can benefit from making use of their agent-like qualities by for example letting them assist us or take over particular tasks, but at the same time we would still be careful about how much we entrust upon them, or be sure to check whether their output is reliable.

There is thus quite a categorical difference between (our treatment of) a system that displays agent-like qualities and one that can truly be seen as an agent. So, rather than doing away with threshold conceptions of agency altogether, we propose to evaluate the agency of artificial systems on the basis of a threshold conception of agency that can minimally account for the fact that humans and higher animals are agents, but that also allows for the possibility of an artificial realization of agency. In the following, we first introduce a theoretical model that can function as such a threshold conception (section 2) and then use this model to determine whether state-of-the-art LLMs can be seen as agents (section 3). We argue that current LLMs are not yet proper agents according to our model, but that elements of these systems suggest ways in which agency might be realizable in an artificial manner. We end the paper with some reflections on what realizing such an artificial agent might and should entail.

## 2. A THEORETICAL MODEL FOR AGENTS

Building on the work of Sarkia (2021), van Lier (2023) identifies four distinct modeling strategies that can be used in complementary fashion to conceptualize artificial agency. Each of these modeling strategies --- Gricean modeling, analogical modeling, theoretical modeling and conceptual modeling --- can be used to answer a different set of questions about a phenomenon like artificial agency. Gricean modeling is the preferred method when one is looking to gain insights about what it would take to build a specific agent, since it allows one, in a stepwise manner, to reconstruct what 'inner' mechanics are most likely to result in the observed behavior of the entity or system that is modeled.[2] If one is rather interested in the way that the phenomenon is similar or different from other phenomena, then analogical modeling is the more useful approach. Where one focuses in Gricean and analogical modeling on (the interrelations of) *individual* phenomena, in theoretical modeling one models the domain of phenomena *as a whole*. With this modeling strategy one can answer demarcation questions like how to differentiate agents from non-agents, or science from pseudoscience. Finally, conceptual modeling can be used to answer questions about the complementability of the models themselves. By reconstructing each of the models built in the other strategies as a conceptual model --- so as a representation of the logical structure of the (sub-)concepts used in the model[3] --- one can see whether or not there is consistency in the way that each of these models refers to the phenomenon in question. In van Lier (2023), these four strategies are combined into one methodological framework, the *Four-Fold Framework,* and it is shown for each of them how they can be used to model artificial agents.

In this paper, we will make use of the third strategy, *theoretical modeling*, to construct our agent model. This is because we are interested in the question of how one can demarcate agents from non-agents when one assumes that these agents do not necessarily need to be alive or mentally endowed (thereby leaving room for the option of an artificial agent). In theoretical modeling, one reasons "about the laws and regularities that are associated with a particular domain of phenomena without detailed reference to either particular entities that populate that domain (as in analogical modeling) or particular mechanisms that maintain those laws and regularities (as in Gricean modeling)" (Sarkia, 2021, §4). In this paper, we demarcate agents from non-agents on the basis of how they are embedded and interact with their environment, and we will therefore model *the laws and regulations that characterize the way that agents are influenced and interact with their environment*.

---
[1] The fact that current LLMs hallucinate, e.g. give us answers that are completely unrelated to the questions we posed, is one indication that they are not agents yet. Another typical example in machine learning are adversarial examples, i.e. cases built from minimal variations of valid inputs which still lead the model to give completely faulty responses.
[2] This modeling method is based on the work of Grice (1974), and has its roots in philosophical psychology, which is also known as the philosophy of mind and action.
[3] Conceptual modeling is based on the model approach that was introduced in the work of Betti & van den Berg (2014, 2016).

There are two advantages to taking a top-down approach like theoretical modeling when debating the potential agency of LLMs. As stated in the introduction, there is a certain resistance in philosophy to attribute agency to artificial systems. One of the reasons for this resistance is that agency is often associated with a form of consciousness and/or a form of biological self-regulation. Both consciousness and mechanisms like autopoiesis seem very difficult to realize in artificial form, and this naturally results in a skeptical attitude towards the possible realization of an artificial agent. By building a theoretical model that models instead the way that agents are dynamically embedded in their respective environments, we are not pre-conditioning agency on being alive or on having some form of consciousness. A second advantage is that we do not have to start from a position where we already assume that LLMs can display agency --- our model neither affirms that there can be artificial agents, nor affirms excludes it as a possibility, since we do not pre-condition agency on being alive or on being conscious. Still, if one accepts that our theoretical model represents the domain of agents, and LLMs fit the model, then one can at least make the argument that agency could be attributed to them.

In the following, we think of humans and most higher animals as paradigmatic examples of agents, and model the way that these agents are dynamically embedded in their environment. We thereby explicitly avoid conditioning these dynamics on them *being alive* or on having some form of *consciousness*.

## 2.1 Framing the model

Just like us, most agent accounts in philosophy have taken (rational) humans and higher animals as the basis for their agent models. However, since we want to widen the agent domain so as to potentially include artificial systems, we need to find some common ground between natural and artificial systems. We have chosen to focus on *autonomy*, meaning that our model attempts to capture *agentive autonomy*. There are two motivations for this focus. First, in philosophy, autonomy is seen as a fundamental characteristic of (specific kinds of) agents. In *the Routledge Handbook of Philosophy of Agency,* for example, Ferrero (2022) lists autonomy as one of the characteristics of what he calls "full-blooded agency" (p. 8). By modeling agentive autonomy, our account would thus still heed the overall consensus about what it means to be an agent in philosophy. A second motivation is that there is a general drive in AI-research to develop autonomous systems. There is thus at least the belief among computer scientists that something like artificial autonomy can be created.

It must be noted, however, that a distinction can be made between the drive of these researchers to develop systems that are 'autonomous' in the sense of them being *self-driven* versus the aim to develop a system capable of a form of autonomy that is more like ours. In Canty et al. (2023), for example, agentive autonomy is described as "adaptive operation" (see the top right of table 1, p. 1260). A self-driving laboratory is autonomous in this sense. Self-driving laboratories are robotic platforms, able to conduct experiments autonomously, to which an AI-system has been added. This AI-system is able to learn how to generate hypotheses, how to design experiments for the robotic platform to test these hypotheses, and how to use the results of these tests to generate new hypotheses.

It is thus able to adapt its operations to what it has learned, enabling it to continue functioning for an extended period of time without needing human feedback. It can be said that such systems are autonomous in the sense that they can 'drive themselves' and this is indeed a feature that characterizes agents as well.

However, as noted in the introduction, *what* an agent ends up doing is influenced by what the agent, as an individual system, has experienced over time. An agent's actions are therefore *authentic* as well: they are "not the product of external manipulative or distorting influences" (Prunkl 2023, p. 101). The distinction made between autonomy as being self-driven versus autonomy as being authentic hints at a similar kind of division in philosophy between two uses of the term autonomy. In its biology-based understanding, autonomy is seen as a form of autopoiesis or self-regulation (see Varela et al. 1991). We think of autonomy as being self-driven and autonomy as self-regulation as similar in that they both seem to primarily relate to the ability of a system 'to continue going by itself', whether this is in the form of surviving (organisms) or not needing human feedback (self-driving systems).[4] Both forms of autonomy are rather abstract in that they do not tell us much about the system as an individual system --- about the system as an *agent*. This is because the how and what of what the self-driven system ends up doing is still prescribed by humans, and the self-regulatory system is directed in how it attempts to survive by its particular nature. In the Kantian understanding of autonomy, on the other hand, autonomy is seen as a form of self-governance (see Formosa 2022, section 2). This is a very advanced form of autonomy since it presumes that agents are able to govern their own behavior with self-made rules. This kind of autonomy requires the agent to have some form of consciousness and it therefore does not meet our aims in building our model. Only the notion of autonomy as being authentic, then, meets our conditions in that it includes the agent as an individual system, without requiring explicitly that this agent has some form of consciousness as well. Since we stated in the introduction of this section that in modeling agentive autonomy we want to avoid conditioning this autonomy on being alive or on being conscious, we will present a model of agentive autonomy where this autonomy is understood as being *authentic* in some way.

We thus aim at constructing a model of autonomy that can engage with state-of-the-art-AI systems, while still staying true to philosophical perspectives on what it means to be an agent. We thereby start from the assumption that the domain of autonomous systems is larger than the domain of systems that philosophers in mind, action and philosophy would attribute agency to, since those philosophers associate agency with autonomy as being authentic or being self-governing. We now hold that *all* agents display a particular kind of autonomy, one that can be meaningfully distinguished from that of other autonomous systems in that what the agent ends up doing is authentic.[5] We further hold that agents are authentic in that what they end up doing relates back to what those agents, as individual systems, have experienced and learned over time. In constructing what this agentive autonomy entails, we take inspiration from the works of Steward (2012) and Walsh (2015). Steward (2012) defends a view that attributes agency to most higher animals. Agents, according to her, are "entities that things can be up to" (p. 25). However, this does not mean that anything can be up to an agent, since "it is utterly undeniable that all animal agency takes place within a framework which constrains, sometimes very tightly, what can be conceived as a real option for that animal" (p. 20). We take away from this account that what is up to an agent is what it does next, but that what it can do next is constrained by a particular framework. Walsh (2015) defines agents as goal-directed systems that, because of their *adaptive repertoire*, can experience their conditions as things that afford opportunities for, or impediments to, the pursuit of their *goals* (p. 163). We will integrate each of these key concepts --- adaptive repertoire and goal --- in our model, since they help define both the drive of the agent and its particular embeddedness in its environment. Having these preliminaries in place, we now turn to our theoretical model.

---

[4] Where we would say that the systems that philosophers of mind, action, and agency refer to as agents are *also* capable of this form of autonomy, we would not similarly say that every system capable of autonomy as being self-regulatory or being self-driven is also capable of a more advanced form of autonomy such as being authentic or being self-governing. As an example of the self-regulatory case, we would say that uni-celled organisms are self-regulatory but not *authentic*, since *how* they react to external impressions like light seems to be fully dependent on the dispositions of their kind and not on what they experience or learn as individuals over time.

[5] A subgroup of these agents --- humans --- can be said to be capable of self-governance as well.

## 2.2 A theoretical model of agentive autonomy

Let us start with the basics. Agents are autonomous systems in the sense that things can be up to them, and we claim that what can be up to them is *always* influenced by a particular and dynamic framework of factors. We thus characterize the autonomy of agents by the way that it is *restricted* by this framework. In most agent accounts in philosophy, especially in free will and determinism debates, the focus lies instead on the way in which agents are *free*.[6] What one tries to answer in such debates is *how* things can be up to the agent --- how they are 'free' --- and this seems to depend largely on what inner processes result in the authentic behavior that we observe, which might be different for each type of agent. For answering these types of questions, then, Gricean modeling seems to be the better method, since it can be used to reconstruct what 'inner' processes are most likely to result in the observed behavior.

In this paper, we will leave these kinds of questions to the Gricean modelers and build instead a theoretical model of the way that agents are dynamically embedded in their respective environments. As stated before, we hold that this embeddedness can be characterized by the fact that what can be up to an agent is *always* influenced by a dynamic framework. This framework consists of three factors --- the agent's *accessible history*, its *adaptive repertoire* and its *external environment*. The latter two factors play a role in most agent accounts in philosophy: agents can act and how they can do so depends on what their particular realization allows them to do (their adaptive repertoire)[7] and what is possible in the environment they encounter. We now claim that what agents can do is *also* influenced by their accessible history --- the set of things that the agent, as an individual system, has learned and achieved over time.

These dimensions of the framework are closely intertwined and sometimes overlap, but one can still make a meaningful distinction between them. The agent uses its accessible history to determine the way in which it will use its adaptive repertoire to interact with its environment and this accessible history changes because of these interactions. The adaptive repertoire is the set of possible ways in which the agent is able to interact with its environment (see Walsh, 2015, p. 211). It is adaptive in the sense that it can be enriched by what the environment offers to the agent and by the agent learning to use elements of its repertoire in new ways. The environment itself consists of everything that is external to the agent, meaning anything that is not part of the agent itself. By interacting with it, the agent can alter its environment, which might result in new options for interaction and learning. The framework thus influences what the agent can do and the agent's actions, in turn, alter the way that this framework factors into what the agent can do.

---

[6] An example of such a philosophical discussion is the work of Steward (2012). In her book, Steward defends *Agency Incompatibilism*, the view that "agency itself is incompatible with determinism" (p. 1). Most agency incompatibilists hold that for agency (read: *human* agency) to be possible, the world cannot be fully determined. Since humans exist, and agency is thus possible, this means for them that universal determinism can thus not be true. Here universal determinism is understood as the claim that "whatever happens anywhere in the universe (every state of affairs) is necessitated by prior events and circumstances, in combination with the laws of nature" (Steward, 2012, p. 9). Steward argues in her book that most animals, if not all, have agency and that even the simplest of these agents are incompatible with a fully determined world. Since there are agents, universal determinism is false, meaning that, for her, agents are to an extent *free* in what can be up to them. Where we do not take a stance in this paper about whether agents can truly be said to be 'free' in what can be up to them, we are sympathetic towards Steward's view.

[7] We borrow this term from Walsh (2015). We deliberately chose not to use the term 'embodiment' here, which might exclude most artificial systems without cause. The agent's adaptive repertoire derives *from* its particular realization, which can be a form of embodiment, but does not necessarily need to be so.

All agents are thus influenced in what they can do by a dynamic *combination* of these three factors. We can now make a distinction between *types* of agents on the basis of how each of the three factors can pan out. We humans, for example, are able to act within complex environments like cultures. Our opposable thumbs have allowed us to add tool-use --- like writing --- to our repertoire, and our writing and communication abilities have made it possible for us to not only make use of our own accessible history, but that of others as well. Bees can also share some of their accessible history with other bees --- like where they have found pollen --- which allows them to work together with others. However, their repertoire is more restricted than ours, which influences how much they can share. The way that each of the factors can pan out thus depends on the *type* of agent that we are dealing with. All agents are thus autonomous in that things can be up to them, but what can be up to an agent might differ from one agent to the next. This is because the influence of the framework on the actions of the agent is dependent on the individual behavior of an agent of a specific type. The fact that agents are influenced by this framework of factors and that these factors unfold quite similarly for the same type of agents, explains why (especially complex) agents have been able to work together and build complex societies. Even though we cannot fully predict the individual behavior of an agent since each of them has followed their own unique trajectory over time leading to small differences in responsiveness from one agent to the next, we can still anticipate the most likely behavior agents are likely to display on the basis of what we know about their history and the type of agent that they are.

We have chosen to refer to the form of autonomy that the agents in our model display as an *indeterminate determinability*, a term that we borrow from the work of Husserl (1989). In the second book of his *Ideas*, Husserl analyzes the way that we are able to understand other persons. He states:

> A person has, in the broadest sense, a typical character and properties of character. Everything a person lives through enlarges the framework of his pregivennesses, can emerge again in memory whether clearly or obscurely, can affect the Ego and motivate actions. But even without memory, it determines the future content of lived experience according to the laws of the new formation of apperceptions and associations. The person is formed through 'experience'. (Husserl, 1989, p. 283)

Each person has a character, which is defined by Husserl as a person's "style of life in affection and action, with regard to the way he has of being motivated by such and such circumstances" (ibid.). This character forms a person's 'framework of pregivenness'.

Persons can have similar characters, and thus a similar framework of pregivenness. What persons differ in is the way that this framework is enriched by their lived experiences over time. Each person will encounter different situations and, depending on their character, experience these situations in various ways (think for example of how the same situation appears to an optimistic vs. a pessimistic kind of person). Husserl now holds that:

> one can to a certain extent expect how a man will behave in a given case if one has correctly apperceived him in his person, in his style. The expectation is generally not plain and clear; it has its apperceptive horizon of *indeterminate determinability* within an intentional framework that circumscribes it, and it concerns precisely one of the modes of behavior which corresponds to the style. (Husserl, 1989, p. 283, italics are ours)

We can thus form a general idea of what a person will do next on the basis of its character. However, this character only determines the range (horizon) of possible behaviors, since to truly grasp the motivations of this other person, we must literally place ourselves in their shoes:

> the subject is not a mere unity of experience, although experience and universal type play an essential role, and it is important that this be brought out and clarified. I put myself in the place of the other subject, and by empathy I grasp what motivates him and how strongly it does so, with what power. (...) I secure these motivations by placing myself in his situation, his level of education, his development as a youth, etc., and to do so *I must needs share in that situation*; I not only empathize with his thinking, his feeling, and his action, but I must also *follow* him in them. (Husserl, 1989, p. 287)

Only by going through all the experiences of the individual kind, then, can we fully get to know and understand what drives the other person.

Our understanding of how the other person will behave can thus be characterized as an *indeterminate determinability* --- it can be determined up to a certain extent on the basis of that person's character, but the remainder is indeterminate since we do not have full access to all aspects and experiences of that person. We think that this distinction between a person's behavior being predictable on the basis of its character but unpredictable on the basis of its individual kind fits quite well with our theoretical model. After all, in our theoretical model, agents of the same type will be capable of a range of behaviors that is characteristic of agents of that type. What this behavior will be at a certain point in time can be estimated by us if we are familiar with that type of agent: it is predictable. However, since we can neither know nor foresee all the myriad things that an individual agent experiences over time, its behavior is never fully predictable to us --- it is an indeterminate determinability.

Agents are thus autonomous systems that things can be up to and *what* can be up to them is always influenced by a particular framework of factors that directly relate to what that individual has experienced and learned over time. This still leaves the question of *how* things can be up to an agent. For any system, to start moving on its own, there has to be something that sets it in motion. This can either be an external cause, or it can be something intrinsic to the system itself. Without a kind of driving force behind it, there is nothing that makes the system move. This driving force, as we see it, is a goal or task that is set *for* the system by its designer (whether this is a programmer or, metaphorically, mother nature). All autonomous systems are goal-directed in this sense. An autonomous system can learn to "mobilise its resources [its repertoire] in a way that is appropriate to the pursuit of its goals, by exploiting the opportunities, or by ameliorating the impediments'' (Walsch, 2015, p. 217). It can thus learn to discern what features in its environment afford opportunities for, or impediments to pursuing its goal, given the repertoire that it has.

Where every autonomous system is goal-directed in the sense that it is driven by a pre-given set of goals, we claim that agents are *purposive* systems in the sense that they can develop additional goals, aims, intentions, or plans. The agent, as a token-system, is able to develop a *purpose* --- a set of *new* goals, aims and/or tasks --- during the pursuit of their pre-given set of goals. This purpose lends additional *authenticity* to the actions of the agent, in that its development is a direct result of how the agent is pursuing its pre-given set of goals while being influenced by its particular framework. A capuchin monkey, for example, learns from its peers that it can find food (pre-given goal) by cracking things open with a stone. Having learned this skill, it can now develop new goals in trying to crack open any kind of object that it encounters, not only those that it has seen its peers try to open. In trying to catch prey (pre-given goal), some octopuses have learned that they can benefit from hunting together with certain fish species (Sampaio *et al.*, 2021). Learning this, they might develop new goals like trying to find fish when going out to hunt again. These new goals, in turn, might influence how their framework will continue to pan out. The purpose that a token-agent develops therefore depends on the way that its framework influences its actions, and this purpose's cultivation, in turn, depends on how the agent's framework will continue to develop and influence its actions.

In our theoretical model, agents are thus autonomous systems whose autonomy can be best described as an *indeterminate determinability*. We can predict quite well what a certain type of agent will do next when we know about its pre-given goals and the combination of factors that usually influence what can be up to the agent of that type. However, we can not fully predict the agent's way of functioning because each individual agent follows its own unique trajectory that can lead it to develop new goals. Having presented our model, we will, in the next section, assess whether current state-of-the-art LLM systems fit the model.

# 3. LLMS AS ARTIFICIAL AGENTS?

"Large language model" (LLM) is a very general term used to describe a plethora of machine learning models able to generate text in various scenarios. Such models consist of a deep neural network, usually based on the Transformer architecture (see Vaswani *et al.*, 2017), containing billions of training parameters (compared to the few millions commonly used for other machine learning tasks). The training of such base LLMs[8] can vary slightly from one to the other, but their core goal is always to predict the word that follows a given input sequence, usually referred to as the prompt. For instance, the model learns that the sequence '*A cat is ...*' should be followed by '*... an animal*'. However, since these models are probabilistic, the answer '*... cute*' would be as good as the previous one. What the model learns is the probability of an answer appearing after an input sequence. So in our example, the model might learn that '*... an animal*' has 90% chance of appearing, while '*... cute*' only has a 10% chance (when no extra context is given to the input sequence).

Even though these base LLMs are very powerful, researchers quickly realized that the answers they produced differed in language and content from what a human would have answered in a particular context. This misalignment hinders the performance of base LLMs in many applications. One solution that researchers have found is to fine-tune the base LLM (i.e. retrain it for a short time) for use-cases. Chatbots, for example, have to be able to give answers that are relevant to the question being asked. To ensure that this is indeed the case, researchers have fine-tuned one of the most prominent base LLM, GPT-4 (Achiam *et al.*, 2023), to output sentences that humans would consider to be good in a certain context (this is known as reinforcement learning from human feedback (Christiano *et al.*, 2017 ; Lambert *et al.*, 2022)).

Due to their size, training base LLMs is very costly, and only few institutions and companies have the resources to do so. Once they are trained, however, their deployment is much cheaper. Nonetheless, even without any further training, researchers have realized that LLMs can *learn* new information just from text, turning them into *zero-shot learners*. For instance, they can learn to perform new tasks by means of some input text that prompts the model before any other prompt that we input to the model (i.e. a *pre-prompt*). For instance, let's consider that an LLM has no previous knowledge of what a multiplication is. We can create a pre-prompt that says "*Whenever you are asked to do a multiplication between two numbers A and B, add A to itself B times*". Now, each time we input two numbers to the LLM, it first sees the pre-prompt and then uses that information to perform the operation. Such a simple feature, which does not entail any further training, can be used in much more complex scenarios, paving the way for the creation of complex LLM architectures that go beyond text prediction.

---

[8] Following the literature, we use the term 'base LLM' to refer to LLMs that are not fine-tuned yet for a specific task.

Perhaps the most promising of these architectures are those that are referred to as *autonomous* (Boiko *et al.*, 2023) or *self-managing* (Firat & Kuleli, 2023) LLM 'agents' [9]. These LLMs consist of a base LLM (like e.g. GPT-4 (Achiam *et al.*, 2023), LLaMa (Touvron *et al.*, 2023), Mistral 7B (Jiang *et al.*, 2023) or Gemini (Team Gemini, 2023), to name a few) that has access to various modules with which it can perform 'actions' that go beyond text generation. For instance, the base LLM can be combined with a web search module to find new information or with a module that allows them to control a robotic arm to perform a chemical experiment (like e.g. the Coscientist of Boiko *et al.* (2023)), or they can be combined with other machine learning models like an image generator (such as Dall-E (Ramesh *et al.*, 2021)). All the information about the 'actions' the LLM can take, i.e. the modules it can use and their functioning, is given as a pre-prompt to the system. This pre-prompt thus contains information about what predefined commands the LLM can use to interact with each of the modules. The model can then use plain text containing these predefined commands to interact with these modules (e.g. the Coscientist in Boiko *et al.* used the text 'GOOGLE suzuki reaction conditions optimal' to activate the web search module and find information about a chemical reaction it later produced in a real laboratory) and use the output of these modules to achieve the task set for it by the initial human prompt (to return to the previous example, Boiko *et al.*'s Coscientist initial prompt was 'You need to perform a Suzuki reaction using the available reagents').

The use of pre-prompts thus makes it possible for researchers to instruct LLMs on how to gain access to, and use, various modules. This has resulted in artificial systems that are capable of quite complex and independent ways of functioning. Where there still needs to be an initial prompt to set the system in motion, it has been shown in works like that of Park *et al.* (2023) and Boiko *et al.* (2023) that little to no human intervention is needed for the completion of the tasks that we set these advanced LLM 'agents'. One could thus say that these 'agents' are capable of a form of autonomous functioning: they are able to learn (during training and fine-tuning, and with the help of pre-prompting) and use what they have learned to continue to function without human intervention for a prolonged period of time (until completion of the task). However, calling this a form of *agentive* autonomy seems, as of yet, a bridge too far. At the moment, LLM 'agents' are fine-tuned for a particular context and instructed on how to use particular modules. They are prepared for use-cases, for performing particular tasks like designing an experiment or holding a believable dialogue. They are meant to function as well, if not better, as we would in a task. For this, they only need to be a *type* of LLM 'agent', one that can proxy a number of human behaviors.

If we want to explore whether these LLMs can be more than our proxies, however, we will have to think of a way in which we can create an *individual LLM agent*: one whose actions and goals are influenced by a dynamic framework of factors which, in turn, is influenced by what the individual LLM agent experiences and learns over time. Of course, this is already happening to an extent since chatbots are able to learn from their interactions with humans and they change their interactions with us on the basis of this. However, these changes are closely monitored so as not to result in unwanted (e.g. racist, sexist, biased or simply wrong) output (see Achiam *et al.* 2023). What is more, these chatbots are trained to proxy human forms of discourse, so where they can thus certainly learn from interacting with us, the *how* and *what* that they should learn is still prescribed by us humans.

---

[9] The quotation marks here are meant to differentiate between the philosophical use of the term agent and its use in AI and computer science. In this paper, we use quotation marks ('agent', 'action') to indicate that we use a term in the latter sense.

The question therefore remains what kind of agent architecture can realize an individual LLM agent. A possible answer can be found in the work of Park *et al.* (2023). The generative agents introduced by these authors fit our theoretical model in the sense that they can make plans and that what they do is based on the current state that they are in as well as their motivations and past individual experiences. These actions, in turn, change their environment and are added to their individual history, which will then again influence any new actions and plans of the generative agent. This agent architecture thus presents a possible way to realize an individual LLM agent. In the simulation itself, which is a closed-world, these generative agents definitely fit our model and a case can therefore be made that, *within the simulation*, they can be thought of as genuine agents.[10] However, one has to take into account that already within this simulation, problems relating to memory capacity arise, such as the generative agents retrieving the wrong memory, or only part of the memory, hallucinating events, or them embellishing their knowledge with the wrong facts from the "world knowledge encoded in the language model used to generate their responses" (Park *et al.*, 2023, §6.5.2). It is therefore still very much a question whether this agent architecture would work as well when exposed to an open world like ours. In thinking about how we can make the transfer from simulation to the real world, we look at the Coscientist that is introduced in Boiko *et al.* (2023). We argue that where this Coscientist is still more like a tool than an agent, its architecture suggests a way in which an artificial agent can be realized in a non-simulated environment. In the following, we bring these two works together and theorize what combination of their architectures can bring us closer to realizing an individual LLM agent that fits our theoretical model.

## 3.1 LLM agents

Before discussing the work of Park *et al.* (2023), we briefly summarize here our theoretical model as discussed in section 2. We stated that agents are systems that things can be up to. What can be up to these agents is always influenced by a dynamic framework of factors that include the agent's accessible history, its adaptive repertoire and its external environment. This framework is dynamic in that it is influenced by the agent's actions and goals and, in turn, can influence these actions and goals. The agent's *type* determines the range in which the factors of the framework can pan out, its set of pre-given goals and the kind of new goals it is likely to form. However, it is the *individual* agent's trajectory through an environment over time that determines how the framework concretely evolves and what specific goals the agent will form. For an individual LLM agent, therefore, there needs to be a way in which what the system does and what goals it forms can be influenced by this framework and, in turn, these goals and actions can influence how the framework develops.

In the work of Park *et al.* (2023), the authors introduce what they call *generative agents* that inhabit a Sandbox world. Even though these generative agents only simulate human behavior, each of them is able to display consistent behavior over a longer period of time and we believe that this is partly due to the fact that their 'actions' and 'goals' are influenced by a framework of factors that bears similarity to the theoretical model that we just described. To better understand this connection, we first want to go briefly over the basis of the system. Every generative agent controls a bot living in the simulated sandbox Smallville. They are pre-prompted with all necessary information needed to properly interact with such an environment, e.g. how to move, use the different objects available, talk to other bots, and what is characteristic behavior for each of the bots. All of these interactions happen via natural language, which is why the generative agents only need to be equipped with an LLM (ChatGPT in this case). Humans can interact with the bots in various ways: by controlling other bots of the sandbox, by changing parts of the environment, or as an "inner voice", i.e. changing or extending the pre-prompts the generative agents have access to.

---

[10] Park *et al.* (2023), however, do not maintain that their generative agents display genuine agency: "the behaviors of our agents, akin to animated Disney characters, aim to create a sense of believability, but they do not imply genuine agency" (footnote 1).

In Park *et al.* (2023), each 'action' of the generative agents is always influenced by a dynamic framework of factors that is quite similar to the one described in our theoretical model. As commented, every generative agent starts off with a pre-prompt that is called its *seed memory*: "one paragraph of natural language description to depict each agent's identity, including their occupation and relationship with other agents" (Park *et al.*, 2023, §3.1). This seed memory tells us something about the *type* of generative agent that we are dealing with. It includes the generative agent's pre-given goals (John Lin, for example, "loves to help people" and "loves his family" (§3.1), while Eddy Lin "loves to explore different musical styles and is always looking for ways to expand his knowledge" (§4.3)), its day-to-day occupation and relation to other bots (John Lin is a pharmacist and his son, Eddy Lin, is a musician) and its innate character traits (Eddy Lin is "friendly, outgoing, hospitable" (§4.3), while his dad is "patient, kind, organized"[11]). The seed memory thus provides us with a general idea of what kind of *accessible history* these generative agents will have. It also tells us something about their *repertoire*: John Lin's 'actions' will relate to his work or his family, while Eddy will do something with music. This initial seed memory is saved to the generative agent's memory stream and determines together with the current state that the generative agent is in (its environment) the first 'action' of the bot once the simulation starts. 'Actions' are therefore influenced from the start by a framework of factors that is similar to the one in our theoretical model (accessible history, repertoire, and environment).

During the simulation, the 'actions' and 'goals' of the *individual* bots influence how their framework pans out and this *dynamic* framework, in turn, influences the 'actions' and 'goals' of the individual bots. To see what we mean, let us start with the bot's 'actions'. Every 'action' a bot performs and the effect this 'action' has on its environment are saved in the bot's memory stream, which is effectively appended to its original seed memory. This original seed memory thus evolves over time, based on the 'actions' the *individual* bot performs.[12] Before performing any new 'action', the bot accounts for its environment and its current memory stream. To filter the memory and avoid irrelevant information, the bots are equipped with a retrieval function that takes as input their current situation and extracts the relevant pieces of memory. The bot then uses such refined memory to plan its next move. Additionally, the bots are able to *reflect* on the basis of their memory. This process, which is automatic but only takes place a few times "per simulated day", allows the bot to see its recent memory stream and summarize its behavior. For example, if a bot has performed various actions related to music, a possible reflection would be: "I like music". Such reflections are added into the memory stream, improving the bot's understanding of its own behavior and guiding its next 'actions'. The individual bot's 'actions' therefore influence how its memory stream (accessible history) is built up, and this memory stream together with the bot's current state (its environment) influences what actions (with its repertoire) the bot is able to perform.

A similar kind of construction influences what 'goals' the individual generative agent is able to form and how these 'goals' in turn influence the 'actions' of the bot. Each generative agent starts the day with a plan that is based on "the agent's summary description (e.g., name, traits, and a summary of their recent experiences) and a summary of their previous day" (§4.3). This plan is thus based on the bot's memory stream (its accessible history). The plan (or goal) can change, however, based on what the bot encounters in its environment during the day. In the simulation, for example, Isabella Rodriguez is prompted to organize a Valentine's party. She invites nine of the other bots, of which three invite additional bots (§7). Each of the invited bots did not have the Valentine's party in their plan for that day, but after receiving an invitation, and based on whether they accepted it (some did not), their plans changed. These bots are now able to develop 'new plans' (goals) and what plans they form is influenced by their individual history and what happens in their environment. Once the day is over, the plans are saved in the memory stream from where they can influence later plans and/or actions.

---

[11] https://github.com/joonspk-research/generative_agents/blob/main/environment/ frontend_server/storage/base_the_ville_n25/personas/ John%20Lin/bootstrap_memory/scratch.json

[12] An example is when the bot Sam Moore runs into the bot Latoya Williams. According to Sam's seed memory, he does not know Latoya. However, after talking to her, this experience is saved in Sam's memory stream and Latoya is now someone Sam knows. She has become part of his accessible history. When he later runs into her again, he remembers their conversation and this determines how he will interact with her (§3.4.2).

It is because these generative agents' 'actions' and 'goals' are always influenced by this specific framework of factors (history, repertoire, environment), and vice versa, that they become believable agents. This is proven by the fact that as soon as one of these factors no longer plays a role, the illusion is shattered. Park *et al.* show that their generative agents are less believable as soon as they cannot reflect (accessible history) or cannot make plans (goals), or do both (§6). Overall, then, we learn from Park *et al.*'s work three things. First, it shows that, at least in simulation, our theoretical model can function well as an agent model. Secondly, it introduces an agent architecture that, when run in a dynamic environment with which the 'agent' can interact, seems to provide a way to realize accessible history and goal development in an artificial manner. Finally, it makes clear that to go from believable agency to genuine agency, we need to focus on what kind of repertoire allows systems to interact with the physical world.

This last point needs to be elaborated on further. A simulated environment that is as complex as Smallville is a good initial training ground, especially if we want to teach a LLM 'agent' a particular set of skills. But what if we want the system to be able to go beyond what we know? What if we want to use artificial systems to create something new, something unexpected, to develop skills and knowledge that we cannot fully anticipate? It seems that to accomplish this a simulated environment is too limiting since all the effects of every possible action are already determined beforehand. If we want artificial systems to become more than our proxies, we need to expose them to the same kind of complex environments that we are exposed to. For this, though, the individual LLM agent needs a repertoire that it can use in an actual environment. To see what shape such a repertoire can take in an artificial setting, we turn to the work of Boiko *et al.* (2023).

As briefly explained above, Boiko *et al.* present Coscientist, a LLM powered bot, that is able to interact with different modules that autonomously design and perform chemical experiments. At the center of Coscientist is an LLM (GPT-4 in the paper), that controls via plain text four different modules: a web searcher and a documentation reader (both built with their own specific LLMs), a Python Code executor and an automation module to direct different devices of a real chemistry laboratory. The central LLM is pre-prompted with the necessary information about the modules, and it is in charge of the general functioning of the Coscientist: sending the right information to each module as well as reading the outputs that each of these feed. The four modules, together with the power of the central LLM, enable Coscientist to perform chemical experiments autonomously after some initial input prompt, such as "*... perform the Suzuki and Sonogashira reactions using the available reagents…*". Given such a prompt and the information contained in its pre-prompt, Coscientist assesses how and when to use each module. For instance, it may first use the web module to get information about such a reaction. Then, it will create a Python program that allows it to compute the quantity of reactant that it should use. Next, it will use the documentation module to learn about a heater-shaker apparatus, which it will then use to perform the experiment in the laboratory.

Not every use-case requires the same set of skills. For most tool-use we need to be minimally able to hold on to the tool in some way. However, using a hammer requires quite different hand-movements than using a screw-driver. The minimal requirements for the use of both are the same (hands with opposable thumbs that can hold on to tools long enough to apply force when necessary), but to learn to use each tool we need to be taught something different. One can think of modules as providing the necessary knowledge to be able to function well in a particular use-case. For looking up something on the internet the Coscientist needs a different set of skills than for doing experiments with physical world hardware. One can thus compare these modules with the tools (or embodiment) that we use to intervene in our environment. They allow us to get to unfamiliar places, to learn new things, to change the environment and observe what happens. We do not use all tools all the time as this would be impractical and cost us unnecessary energy (only imagine having to hold on to our entire toolbox continuously). The same goes for the modules that the Coscientist makes use of. It accesses and uses the module that it needs, takes from it what it requires and then uses another module to use the information it has gotten from the previous module.

We can now say that the Coscientist has a repertoire that consists of the technologies that it can activate through a particular module. These technologies can affect the online environment, but the offline environment as well, for example when it activates technologies like physical world hardware. An individual LLM agent could gain similar access to a dynamic and complex environment with the help of modules. In the case of the Coscientist, however, its repertoire is not adaptive, since it can only use four modules (web searcher, code execution, docs searcher, automation). What is more, it can in principle retain the information it gains through the use of the modules to carry it over to the next prompt, but since the Coscientist is used to solve the prompts it is given, there is no need for it to retain more than is necessary for executing a specific prompt. What the Coscientist is missing, then, is an accessible history. An individual LLM agent should be influenced in what it does and the goals that it forms by a framework that consist of *all* three factors --- its accessible history, its adaptive repertoire and its external environment.

An individual LLM agent, as we imagine it, should have a similar architecture as the Coscientist, one that consists of multiple modules that work together. This would provide it with a repertoire with which it can interact with various environments. It should also incorporate an agent architecture that is similar to that of Park *et al.*, to provide it with its own accessible history, thereby making it into an individual system. The Coscientist's memory stream does not go beyond the experiment it runs. This means that, after successfully performing an experiment, the planner's pre-prompting is completely reset to an initial state where only the information of the modules and its main goals are contained. This is based on the consideration that, for any new experiment we demand, all the information should be contained in the web and thus be available to the bot. Since, at the moment, the Coscientist is 'only' meant to perform the chemical research we request, there is no need for it to act as an individual system. However, it may happen in the future that such bots are able to produce new knowledge from which they can benefit in posterior rounds, hence benefitting from having an accessible history. Given that Park *et al.*'s generative agents fit our model, we suppose that their agent architecture would be a good fit for realizing an accessible history in an artificial setting. Additionally, we might want to allow LLM agents to generate their own modules or to learn to use new ones, so that they can adapt their repertoire on the basis of their actions and goals.

An individual LLM agent should thus have an architecture like that of the Coscientist, but with the ability to adapt modules, and with an accessible history that is based on the agent architecture of Park *et al*. This accessible history has to start with a memory seed that states the role or function of the LLM agent (research assistant, coach, financial advisor), as well as its goals and drives (discovering something new, but staying within the current framework). To develop its accessible history, this individual agent would have to be able to supplement its seed memory with what it learns from interacting with its environment through its adaptive repertoire (the modules). To make use of an adaptive repertoire, the individual LLM agent would need to be able to learn to use modules in a new way (or even be able to write new ones). We believe that a system that can meet these requirements can and should be considered as a potential realization of agency in an artificial setting. It thus seems that the necessary elements are already there, and that the questions that remain are of a more practical and ethical nature. What will it take for us to actually realize such an artificial agent, and what are the challenges we still face?

# 4. CONCLUSION AND FUTURE OUTLOOK

We are currently witnessing a broad-scale integration of AI technologies in our social institutions. Since we are interacting with these systems in the same manner as we do with agents, it is only natural to ask if we have managed to create systems that are more than mere proxies of genuine agency. To be able to answer this question, however, we need a notion of agent that does not exclude artificial systems from the outset because of reasons like that they do not have consciousness or are not alive. In this work, we therefore present a theoretical model that can be used as a threshold conception for artificial agents. We use the model to show that current LLM 'agents' are not agents yet, but that they do contain elements that fit parts of our model. The generative agents in Park *et al.*, for example, are based on an agent architecture that could function as an accessible history for an artificial system and the modules of the Coscientist in Boiko *et al.* provide a way for LLM 'agents' to interact with their environment and these modules could thus potentially be seen as an artificial system's repertoire.

Since the necessary elements appear to already be there, we even venture to propose what it would take for an artificial system to meet our model and to therefore serve as a potential and realistic realization of an artificial agent. Even though their basic architecture, the Transformer, was conceived only a few years ago, LLMs have surged as a key technology that sees enormous developments everyday. In that sense, we may expect that multiple new LLM bots, fulfilling the conditions proposed in the presented framework, already exist by the publication of this manuscript. The two agent architectures described in this work, for example, are paradigmatic examples of a big family of such LLM 'agents'. Other examples are Chameleon (Lu *et al.*, 2023), a similar system to Coscientist, that is able to use a variety of tools (computer-based in this case) to accomplish a wide range of complex reasoning tasks and the Voyager (Wang *et al.*, 2023), a continuously learning agent set in a Minecraft environment, that is able to acquire new skills by combining existing ones in non-trivial ways. The latter could already be seen as a first step to the automatic conception of new modules by a LLM 'agent'. It is thus quite conceivable that a LLM agent that fits our model will be realized in the foreseeable future.

Still, some challenges persist. For instance, LLMs are currently hindered by their 'attention span', i.e. the amount of words they are able to consider from their memory (i.e. all previous prompts) for creating new output. However, such a bottleneck is a principal avenue of research and will for sure see great developments in the near future. It can even be that strategies such as the reflection function proposed in Park *et al.* will allow LLMs to build their own structured memories, which may reduce their size and improve memory retrieval. Besides the challenge of memory retrieval, it may even be the case that, in the future, natural language itself becomes an unnecessary or suboptimal channel for AI agents. Even though it has been shown to be a powerful tool to create communication between different modules, other, more basic or efficient communication strategies may be developed (e.g. electromagnetic signals). Where these are challenges that computer scientists face, there are further challenges that require all of our attention.

One of these challenges is the *seed memory*. In Park *et al.*, the seed memory functions as the generative agent's identity, its drive, its general direction in 'life'. How many instructions should the seed memory of an individual LLM agent contain, and how explicit should these instructions be? Should we start with an agent whose identity is already clearly spelled out? In our view, this seed memory can play a key role in ensuring that the way that the individual LLM agent makes sense of the world aligns with our own views of the world. It is here that we can make sure that it knows about the proper rules of conduct when taking on its role in our social institutions. However, we also do not want to put too much in this seed memory, as this could potentially stunt the development of these systems as individual agents. The seed memory can have a big effect on how the system develops itself, so it is here that we should direct our focus and where we need input not only from scientists but also from philosophers and government officials.

Since it might not be too long before an artificial system is created that fits our theoretical model and that should therefore at least be considered as a potential agent, there is also the challenge of whether and in what way we want to integrate these artificial systems in our society. If their actions are similarly *authentic* like ours, an instrumental treatment of them might no longer be fitting. However, how much status and power do we really want to give to systems that have the ability to access and process far more information than we ever could? We have already found out that much of the data we feed these systems is biased (and this will probably only perpetuate in this age of fake news), and we should make sure that such computationally powerful systems are not negatively affected when exposed to this data. What is more, we already treat many of our own as second-rate citizens. So before we start preparing for the integration of new members in our society, should we not focus on fixing these injustices first? What is often left out of the discussion, for example, is that there is already an unequal divide of resources and our Western drive to build and integrate increasingly sophisticated systems can only happen by the continuing exploitation of the land and resources of third-world countries (Crawford 2023). What is more, these countries and other minorities are most of the time not even the ones who benefit from all that these technologies have to offer (see Benjamin 2019).

Where the current advancements in computer science and AI are thus fast and the option to create an artificial agent seems to have come within our reach, we are now faced with the question of what it will bring us to create a system capable of an *authentic* form of autonomy. Do the benefits outweigh the costs --- for all? And, given the fact that integrating such systems can have far reaching and negative societal impacts when not done well (like the aforementioned increase of societal inequalities, and the costs to our planet), should we not focus instead on building artificial systems who are autonomous in the sense of being *self-driven*? We can then still use them to help us with tasks that require a computational power and precision that we simply do not possess but could benefit from (e.g. quantum error correction, predicting protein shapes, spotting malignant tumors), while keeping the fail-safes in place that fit the instrumental role that they would play.

If we do however choose to continue to develop artificial agents capable of an authentic form of autonomy, there is work to be done. We not only have to make sure that the information that we expose them to to learn about us and our world aligns with our values rather than our vices, but we also have to start them off with the right motivations. This means on the one hand that we have to become more aware of the biases already present in our data and to actively fight the spread of fake news. It also means determining what *role* we want these systems to take on in our society (assistant, companion, teacher) and, in taking on this role, what *perspective* we want them to reason from (to pursue equality and justice, as well as truth and objectivity). The latter considerations then need to be translated into a seed memory that starts the artificial agent off with a steady basis in its exploration of our world. Current state-of-the-art LLM systems show a lot of promise, but for all of us to benefit from what *they* have to offer, we have to teach them the best that *we* have to offer.

## Acknowledgements

Maud van Lier gratefully acknowledges the support by VolkswagenStiftung Grant Az:97721. Gorka Muñoz-Gil acknowledges funding from the European Union. Views and opinions expressed are however those of the author(s) only and do not necessarily reflect those of the European Union, the European Research Council or the European Research Executive Agency. Neither the European Union nor the granting authorities can be held responsible for them.